\documentclass[letterpaper]{article} % DO NOT CHANGE THIS
\usepackage{aaai23}  % DO NOT CHANGE THIS
\nocopyright
\usepackage{times}  % DO NOT CHANGE THIS
\usepackage{helvet}  % DO NOT CHANGE THIS
\usepackage{courier}  % DO NOT CHANGE THIS
\usepackage[hyphens]{url}  % DO NOT CHANGE THIS
\usepackage{graphicx} % DO NOT CHANGE THIS
\usepackage{natbib}  % DO NOT CHANGE THIS AND DO NOT ADD ANY OPTIONS TO IT
\usepackage{caption} % DO NOT CHANGE THIS AND DO NOT ADD ANY OPTIONS TO IT
\usepackage{subfigure}
\usepackage{algorithm}
\usepackage{algpseudocode}
\usepackage{multirow}
\usepackage{array}
\usepackage{siunitx}
\newcolumntype{P}[1]{>{\centering\arraybackslash}p{#1}}

\usepackage{newfloat}
\usepackage{listings}
\DeclareCaptionStyle{ruled}{labelfont=normalfont,labelsep=colon,strut=off} % DO NOT CHANGE THIS
\floatstyle{ruled}
\newfloat{listing}{tb}{lst}{}
\floatname{listing}{Listing}
\usepackage{amsmath,amsthm,amssymb}
\usepackage{enumitem}
\setlist[itemize]{leftmargin=*, noitemsep}

\usepackage{xcolor, colortbl}

\usepackage[defaultcolor=red]{changes} % \usepackage[final]{changes}
\definechangesauthor[color=red]{DJ}

\title{PPO-UE: Proximal Policy Optimization via Uncertainty-Aware Exploration}
\author {
    Qisheng Zhang\textsuperscript{\rm 1},
    Zhen Guo\textsuperscript{\rm 1},
    Audun J{\o}sang\textsuperscript{\rm 3},
    Lance M. Kaplan\textsuperscript{\rm 4},
    Feng Chen\textsuperscript{\rm 5},
    Dong H. Jeong\textsuperscript{\rm 6},
    Jin-Hee~Cho\textsuperscript{\rm 1}\\
    }
\affiliations {
    \textsuperscript{\rm 1}Department of Computer Science, Virginia Tech, VA, USA; 
    \textsuperscript{\rm 2}2810 Jackson Avenue, NY, USA;\\
    \textsuperscript{\rm 3}University of Oslo, Oslo, Norway;    
    \textsuperscript{\rm 4}DEVCOM Army Research Laboratory, MD, USA; \\   
    \textsuperscript{\rm 5}University of Texas at Dallas, Richardson TX, USA;
    \textsuperscript{\rm 6}University of the District of Columbia, DC, USA \\   
    qishengz19@vt.edu, zguo@vt.edu,  audun.josang@mn.uio.no, lance.m.kaplan.civ@army.mil, feng.chen@utdallas.edu, djeong@udc.edu, jicho@vt.edu
} %, thirdAuthor@affiliation1.com

\begin{document}

\maketitle

\begin{abstract}
Proximal Policy Optimization (PPO) is a highly popular policy-based deep reinforcement learning (DRL) approach. However, we observe that the homogeneous exploration process in PPO could cause an unexpected stability issue in the training phase. To address this issue, we propose PPO-UE, a PPO variant equipped with self-adaptive uncertainty-aware explorations (UEs) based on a \textit{ratio uncertainty level}. The proposed PPO-UE is designed to improve convergence speed and performance with an optimized \textit{ratio uncertainty level}. Through extensive sensitivity analysis by varying the \textit{ratio uncertainty level},  our proposed PPO-UE considerably outperforms the baseline PPO in Roboschool continuous control tasks.
\end{abstract}

\section{Introduction} \label{sec:intro}

Deep reinforcement learning (DRL) is a set of reinforcement learning algorithms equipped with deep neural networks (DNNs).  DRL algorithms can be categorized into two: {\em value-based} and {\em policy-based}.  Value-based approaches include Deep Q-network~\cite{mnih2013playing}, Double Deep Q-Networks~\cite{van2016deep}, and Dueling Deep Q-Networks~\cite{wang2016dueling}. These approaches separate the exploration and exploitation processes with additional rule-based methods. Policy-based approaches include Advantage Actor Critic (A2C) \cite{mnih2016asynchronous}, Policy Gradient \cite{peters2006policy}, Trust Region Policy Optimization (TRPO)~\cite{schulman2015trust}, and Proximal Policy Optimization (PPO)~\cite{schulman2017proximal}. These policy-based DRL approaches learn policy functions that directly map states into actions. This allows the exploration and exploitation processes to be combined by sampling actions based on policies. 

In continuous control problems, value-based approaches cannot effectively learn the optimal action from a value function due to the continuous action space.  Thus, policy-based approaches have been widely used to solve these problems. However, the performance of policy-based approaches largely depends on the sampling process. For instance, PPO's Gaussian action exploration mechanism has a stability issue~\cite{ciosek2019better}. In this work, we aim to refine the exploration mechanism in PPO to balance data exploration and exploitation better.  Specifically, we propose a variant of PPO, called \textit{PPO-UE}, to leverage the uncertainty information in learned policies and improve the overall performance of the original PPO algorithm.

\vspace{1mm}
Via {\em PPO-UE}, we make the following {\bf key contributions}:
\vspace{-3mm}
\begin{enumerate}
\item We give a rigorous theoretical analysis of the sampling techniques used in general policy gradient methods. The analysis shows the stability issue in these techniques.
\item We propose an algorithm called {\em PPO-UE}, which enables uncertainty-aware explorations in the training phase. The uncertainty-aware exploration can intelligently adapt to different policy statuses and provide adaptive, state-dependent exploration strategies.
\item We conduct extensive experiments to investigate the impact of the uncertainty metric in our proposed PPO-UE on learning performance in Roboschool continuous control tasks. The experiment results show that PPO-UE can achieve faster convergence and better performance than the original PPO baseline.
\end{enumerate}

\section{Related Work} \label{sec:related-work}

Recently, much work has been done to refine the PPO algorithm further. \citet{xiao2020fast} focused on the policy iteration process and subtracted a baseline term from the advantage function in the loss function to further improve the learning efficiency of PPO.  In addition, more work has been done in terms of the exploration process in PPO. \citet{zhang2022proximal} improved the exploration efficiency of PPO using a revised reward function with an additional term called {\em uncertain reward}. The uncertain reward aims to give incentives for more explorations. \citet{khoi2021multi} took a similar approach to leverage a technique called \textit{Curiosity Driven Exploration}. Similarly, an additional term, called {\em intrinsic reward signal}, is added to the original reward function to improve the exploration efficiency. Following a similar line of ideas, \citet{liu2022capacity} used an additional reward term called {\em internal reward} to encourage more balanced explorations.  However, instead of improving reward functions, \citet{hamalainen2020ppo} proposed an algorithm called \textit{PPO-CMA} to periodically update the covariance matrix of sample (CMA) distribution along with the policy iterations.

Unlike the works above, we consider self-adaptive, uncertainty-aware explorations without changing the sampling distribution and reward function in PPO. Therefore, our approach can provide easy implementability and parameter-tuning capability during the training phase.

\section{Preliminaries} \label{sec:prelim}

In this work, we consider the on-policy DRL algorithms. Given a policy $\pi_\theta$ parameterized by $\theta$ and a state $s_t$ at time $t$, the agent takes an action $a_t\sim \pi_\theta(a_t|s_t) = \pi_\theta(p_e(\mu))$. Here the policy $\pi_\theta$ includes two components: one is $\mu$ output by the actor neural network, the other is the exploration distribution $p_e$ as a function of $\mu$. In the following text, we would refer to $a_t$ as the action output by $\pi_\theta$ and $\mu$ as the policy mean output by $\pi_\theta$. After sampling the action $a'_t$, the new state $s_{t+1}$, the reward, $r_t$, will be given by the environment. The agent's goal is to find the optimal $\theta$ and corresponding $\pi_\theta$ to maximize the accumulated expected reward $E(\sum_{t=0}^\infty \gamma^t r_t)$ where $\gamma$ is a decay factor.

We consider a continuous environment, which is simulated until a predefined terminal state or a maximum episode length $T_e$ is reached. Then, the agent will update $\theta$ and $\pi_\theta$ periodically with a constant update interval $T_u$.

\subsection{Policy Gradient} \label{subsec:pg}

The policy gradient method~\cite{sutton1999policy} updates the policy gradient with two steps. First, the gradient is estimated by differentiating the following loss function:
\begin{equation}
 L^{PG}(\theta) = E_t(\log\pi_\theta(a_t|s_t)A_t).
 \label{eq:loss1}
\end{equation}
Here, $A_t$ is an advantage function used to estimate the benefit of taking $a_t$, given state $s_t$. Second, this process can lead to forming the following gradient:
\begin{equation}
g = E_t(\nabla\log\pi_\theta(a_t|s_t)A_t).
\end{equation}

\subsection{Proximal Policy Optimization} \label{subsec:ppo}
To improve the training efficiency and stability of the policy gradient method, Trust Region Methods (TRPO)~\cite{schulman2015trust} was proposed with the surrogate objective function under constraints on the policy update size. However, the surrogate objective function of TRPO could not accurately estimate the policy performance. Thus, to refine the objective function and reduce the computational overhead, \citet{schulman2017proximal} proposed Proximal Policy Optimization (PPO) to perform multiple minibatch gradient steps in one policy update iteration. Specifically, PPO is refined to improve the original TRPO in two aspects. First, the clipped surrogate objective is used as the substitution for the original surrogate objective. Second, the adaptive Kullback–Leibler (KL)-divergence penalty coefficient is used rather than a constant penalty. \citet{schulman2017proximal} claim that the clipped surrogate objective outperforms the original surrogate objective. Hence, in this work, we use it in all PPO vs. PPO-UE performance comparative evaluations.

\subsection{Continuous Action Spaces} \label{subsec:cas}

In continuous control problems, the policy gradient algorithms, including PPO, output a desired action rather than an action distribution. Thus, we form the action distribution based on a predefined probability distribution model.  A {\em multivariate normal distribution}, a.k.a. {\em multivariate Gaussian distribution}, is commonly used to sample the new actions based on the action output by the policy. The multivariate Gaussian distribution enables the policy to sample the action $a\sim\mathcal{N}(\mu,\Sigma)$, where $\mu$, also known as the policy mean, is the action output by the actor neural network of policy $\pi_\theta$ and $\Sigma$ is the covariance matrix. In this work, we follow the original PPO algorithm~\cite{schulman2017proximal} and employ a diagonal covariance matrix $\Sigma$. Other than using the Gaussian distribution to sample actions for all states, we adjust the sampling process to be well applicable in the Roboschool problem based on the given state.

\section{Proposed PPO-UE Algorithm}\label{sec:ppo-ue}

\subsection{Problem Analysis}
As mentioned earlier, PPO uses a global Gaussian distribution $\mathcal{N}(\mu,\Sigma)$ to sample actions for all states. The predefined covariance matrix $\Sigma$ is independent of the learned policies. Since we have the PDF of $\mathcal{N}(\mu,\Sigma)$ with
\begin{equation}
f(a)=\frac{1}{\sqrt{(2\pi)^d\lVert\Sigma\rVert}}\exp(-\frac{1}{2}(a-\mu)^T\Sigma^{-1}(a-\mu)),
\end{equation}
we have 
\begin{equation}
\log(f(a))=-\frac{1}{2}(d\log(2\pi)+\log(\lVert\Sigma\rVert)+d_M^2(a,\mathcal{N}(\mu,\Sigma))).
\label{eq:log1}
\end{equation}
Here $d_M$ refers to the Mahalanobis distance of action $a$ from $\mathcal{N}(\mu,\Sigma)$. Suppose $a\in \mathbb{R}^d$, $\Sigma = \mathrm{diag}(\vec{\sigma})$, Eq.~\eqref{eq:log1} can be further simplified to
\begin{equation}
\log(f(a))=-\frac{1}{2}\Big(d\log(2\pi)+\sum_i^d \Big(\log(\sigma_i)+\frac{(a_i-\mu_i)^2}{\sigma_i}\Big)\Big).
\end{equation}
Therefore, minimizing the loss function Eq.~\eqref{eq:loss1} is equivalent to minimizing the following,
\begin{equation}
L^{PG}(\theta) = E_t((\sum_i^d(\log(\sigma_i)+\frac{(a_i-\mu_i)^2}{\sigma_i}))A_t).
\label{eq:loss2}
\end{equation}
This means the gradient of Eq.~\eqref{eq:loss2} points to the direction where the policy fits positive-advantage actions other than negative-advantage actions. However, sampling with a global Gaussian distribution cannot distinguish positive-advantage actions from negative-advantage actions. This destabilizes the update process of policy iterations.

\subsection{Uncertainty-Aware Exploration}\label{subsec:ue}

To mitigate the stability issue caused by the negative-advantage actions, we introduce \textit{Uncertainty-Aware Exploration} (UE) to balance the exploration and exploitation further. For a given state $s$, assume the policy $\pi_{\theta_t}$ outputs the optimal action $a$, then the policy $\pi_{\theta_{t+1}}$ should be trained to output the same optimal action $a$, as the policy mean $\mu$ is trained towards positive-advantage actions. This means we should keep exploiting the action if the action taken is good enough. However, in practice, we cannot evaluate the action optimality from the policy. Thus, we need to estimate it from policy iterations.

\subsubsection{Action Distance Ratio}\label{subsubsec:adr}

We aim to approximate the action optimality from two consecutive policies. For a given state $s_t$ at time step $t$ with two policies $\pi_{\theta_{t-1}}$ and $\pi_{\theta_{t}}$, we denote the actions output by $\pi_{\theta_{t-1}}$ and $\pi_{\theta_t}$ as $a_{t-1}$ and $a_t$, respectively. Then we define the \textit{action distance} $d(s_t) = \lVert a_{t}-a_{t-1}\rVert$. Furthermore, we can derive the \textit{action distance ratio} as $r(s_t) = \frac{\lVert a_{t}-a_{t-1}\rVert}{\lVert a_{t-1}\rVert}$. Note that $r(s_t)$ measures the degree of the policy update with respect to $s_{t-1}$. Thus, $r(s_t)=0$ is the necessary condition for policy convergence in $s_t$.  The smaller $r(s_t)$ indicates higher optimality in local actions which are restricted by $\pi_{\theta_{t-1}}$ and $\pi_{\theta_{t}}$. 

\subsubsection{Exploration Threshold and Ratio Uncertainty}\label{subsubsec:etu}

Since the action distance ratio can estimate the optimality of an action taken, we only enable exploration when the ratio is sufficiently high. We need a global view of these ratios to select an appropriate action distance ratio as the exploration threshold. To this end, we rank all ratios $r(s)$ between the $k$-th and $(k+1)$-th updates in ascending order and select the exploration threshold at the desired ranking. Specifically, we set an \textit{ratio uncertainty level} $U_k\in[0,1]$ and define the \textit{exploration threshold} $\tau_k$ as the ratio with the ranking $\lfloor(1-U_k)L_k\rfloor$. Here, $L_k$ is the ranking length. This means the policy will exploit an action as the policy mean $\mu$ only when its corresponding ratio is smaller than $\tau_k$. Otherwise, the agent will choose to sample a new action. As a special case, the original PPO algorithm~\cite{schulman2017proximal} has a ratio uncertainty level $U_k=1$ and $\tau_k=0$ for any $k$.

\subsection{Algorithm Summary}\label{subsec:algo}

To apply UE to the PPO algorithm, we maintain a copy of the old policy when each policy is updated. We cannot calculate the exploration threshold based on the ongoing policy since the sampling process depends on the exploration threshold. Hence, we calculate the exploration threshold from the previous update and use it for the ongoing sampling process. We describe the details of the proposed PPO-UE in Algorithm~\ref{algo:ppo-ue}.

\begin{algorithm}[ht]
\small{
\caption{\small {\bf PPO-UE}}
\label{algo:ppo-ue}
\begin{algorithmic}[1]
\State{$T \leftarrow$ total training steps}
\State{$T_u \leftarrow$ policy update interval}
\State{$k \leftarrow$ policy update iterations}
\State{$T_e \leftarrow$ maximum episode length}
\State{$U_0 \leftarrow$ initial ratio uncertainty level}
\State{$\tau_0 \leftarrow$ initial exploration threshold}
\State{$\pi_{\theta_0} \leftarrow$ initial policy}
\State{$t_1,k = 0$}
\While{$t_1 < T$}
    \State{$t_2 = 0$}
    \While{$t_2 < T_e$}
    \If{$r(s_{t_2})>\tau_k$}
    \State{Sample a new action with $\mathcal{N}(\mu,\Sigma)$}
    \Else
    \State{Sample an action output by $\pi_{\theta_{t_1}}$}
    \EndIf
    \If{$t_1\equiv 0\ (\textrm{mod}\ T_u)$}
    \State{$k=\frac{t_1}{T_u}$}
    \State{$\tau_k=\tau_k(U_k,L_k)$}
    \State{Update policy $\pi_{\theta_{t_1}}$}
    \EndIf
    \EndWhile
\EndWhile
\end{algorithmic}
}
\end{algorithm}

\begin{table}[ht!]
\centering
\caption{Algorithm Hyper-parameter Setting}
\vspace{-2mm}
\begin{tabular}{|P{1.2cm}|p{4cm}|P{1.8cm}|}
\hline
\text{Param.} & Meaning & Value\\
\hline
$T$ & Total training steps & \num{1e6} \\
\hline
$T_e$ & Maximum episode length for training & 512 \\
\hline
$T'_e$ & Maximum episode length for testing & 2,048 \\
\hline
$T_u$ & Policy update interval & 2,048 \\
\hline
$U_0$ & Initial ratio uncertainty level & 1 \\
\hline
$\tau_0$ & Initial exploration threshold & 0 \\
\hline
$\epsilon$ & Clipping parameter in PPO & 0.2 \\
\hline
$K$ & Number of epochs in PPO & 80\\
\hline
\multirow{2}{*}{$\log(\sqrt{\sigma_i})$ } & Log standard deviation of action distribution in PPO & LinearAnneal $(-0.1, -1.6)$ \\ 
\hline
\end{tabular} 
\label{table:param_default_values}
\vspace{-5mm}
\end{table}

\section{Experiment Setup} \label{sec:ex-setup}

To demonstrate the outperformance of our proposed PPO-UE algorithm in terms of high-dimensional continuous control problems, we trained and tested PPO-UE and the baseline PPO on the OpenAI Gym RoboschoolWalker2d~\cite{brockman2016openai}. In this environment, a 3D humanoid must balance multiple factors to walk optimally. The robot is trained to meet the following three goals: (1) moving as fast as possible; (2) finding the least number of actions to perform a move; (3) maintaining a healthy status. We performed $10$ training runs with different random seeds for a given environment and setting. The only difference of PPO-UE from PPO is the exploration phase.  We use the same hyper-parameters for other components in PPO. We also use a longer horizon $T'_e$ in the testing phase to show the generalizability of trained policies. Table~\ref{table:param_default_values} describes the details of the hyper-parameter setting used in this study.

\subsection{Comparing Schemes}\label{subsec:scheme}

To simplify the parameter tuning process, we use a fixed ratio uncertainty level $U=U_k$ for all iteration $k$. Under this setting, we aim to find an optimal ratio uncertainty level $U$ for maximizing the testing reward. We also conduct a sensitivity analysis of $U$ which is ranged in [$0.8$, $0.9$, $0.96$, $0.98$, $0.99$]. To simplify the notations, we rename the corresponding schemes as $\textrm{PPO-UE}_{0.8}$, $\textrm{PPO-UE}_{0.9}$, $\textrm{PPO-UE}_{0.96}$, $\textrm{PPO-UE}_{0.98}$, and $\textrm{PPO-UE}_{0.99}$.  The baseline PPO is equivalent to PPO-UE with $U=1$.  We have six schemes parameterized by $U$ and conduct their comparative performance analysis.

\subsection{Metrics}\label{subsec:metric}

We use three metrics for our experiments: (1) Training reward $R_{train}$ with horizon $T_e$; (2) Testing reward $R_{test}$ with horizon $T'_e$; (3) Posterior ratio uncertainty level $PU$ given by actual action distance ratio rankings. Specifically, after sampling with exploration threshold $\tau_k$, we obtain the action distance ratio ranking for all samples. Denote the number of samples with an action distance ratio below $\tau_k$ as $L_{low}$, we have $PU = 1-\frac{L_{low}}{L_k}$, where $L_k$ is the ranking length. We propose $PU$ to check the consistency of ratio uncertainty level $U$. We discuss more details in the following section.

\section{Experiment Results \& Analysis} \label{sec:results}

We trained PPO-UE and PPO baseline for $10$ simulation runs. Each simulation run includes a total of $1$ million in training steps.  In the testing phase, the agent chooses the policy mean $\mu$ as the next action instead of random sampling. We evaluate each scheme $100$ times to obtain the testing reward.

\subsection{Training Reward}

\begin{figure}[ht]
\vspace{-3mm}
  \centering
  \subfigure{
    \includegraphics[width=0.45\textwidth]{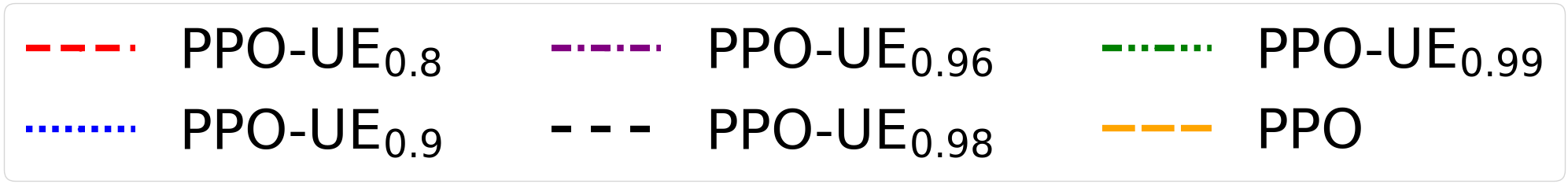}}
   \setcounter{subfigure}{0}
  \subfigure{
    \includegraphics[width=0.45\textwidth]{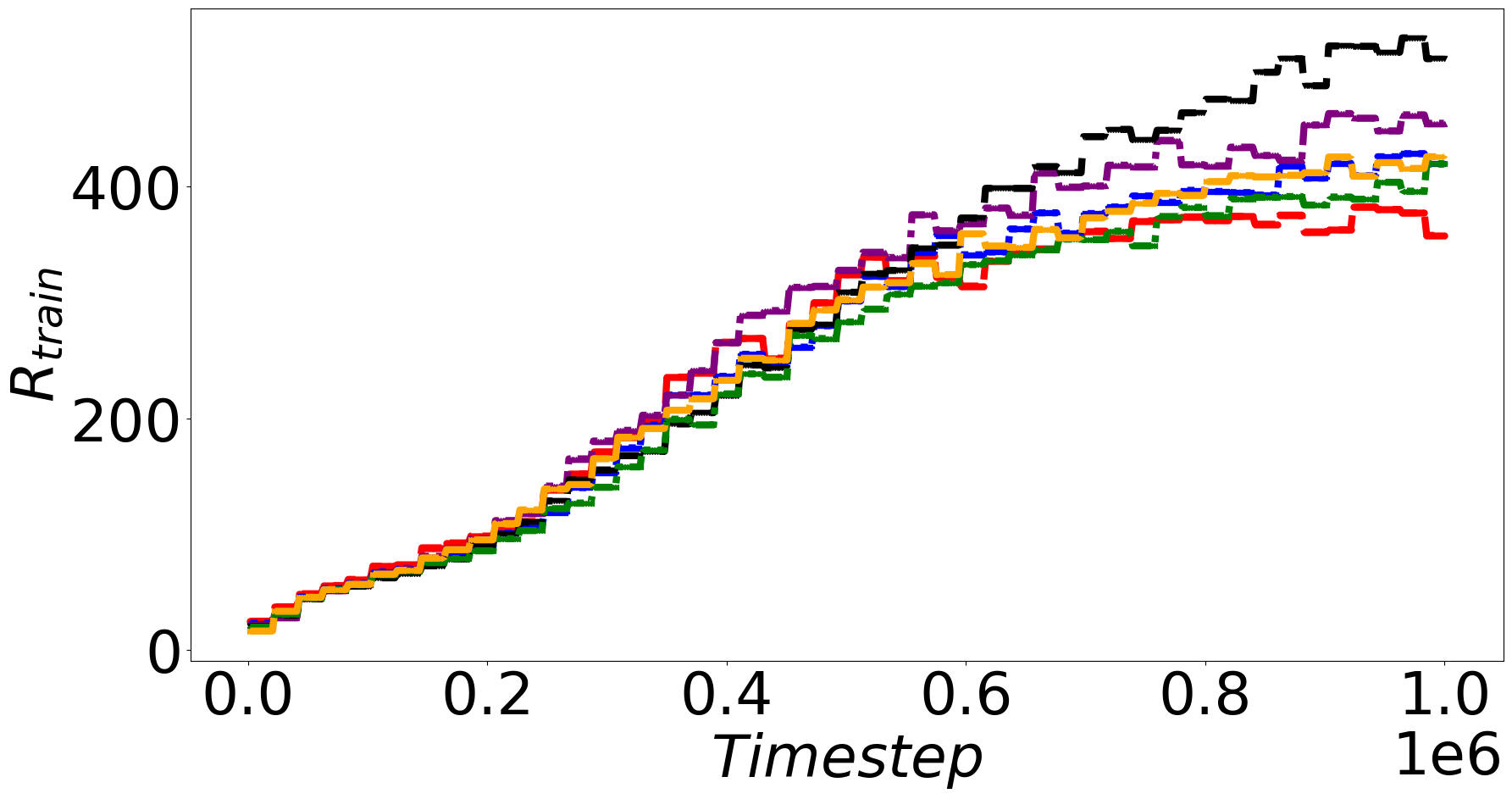}}
    \caption{Training rewards with respect to training time steps.}
\label{fig:train}
\vspace{-3mm}
\end{figure}

Figure~\ref{fig:train} shows the learning curve of the six schemes with respect to training time steps. The overall performance order is: $\textrm{PPO-UE}_{0.98}\geq\textrm{PPO-UE}_{0.96}\geq\textrm{PPO-UE}_{0.9}\approx\textrm{PPO}\geq\textrm{PPO-UE}_{0.99}\geq\textrm{PPO-UE}_{0.8}$. It is clear that $\textrm{PPO-UE}_{0.98}$ performs the best among all schemes after $1$ million training time steps. This implies that there exists an optimal ratio uncertainty level $U$ to maximize the training reward. Furthermore, the convergence speed is also affected by $U$. $\textrm{PPO-UE}_{0.8}$ has the greatest convergence speed but worst performance overall. This is because $U$ is too low to ensure adequate exploration during the training process. Hence, the agent chooses actions based on exploitation excessively and the corresponding policy quickly converges to a sub-optimal solution. In general, a smaller uncertainty level, $U$, indicates a faster convergence.

\subsection{Testing Reward}

\begin{figure}[ht]
\centering
\includegraphics[width=0.45\textwidth]{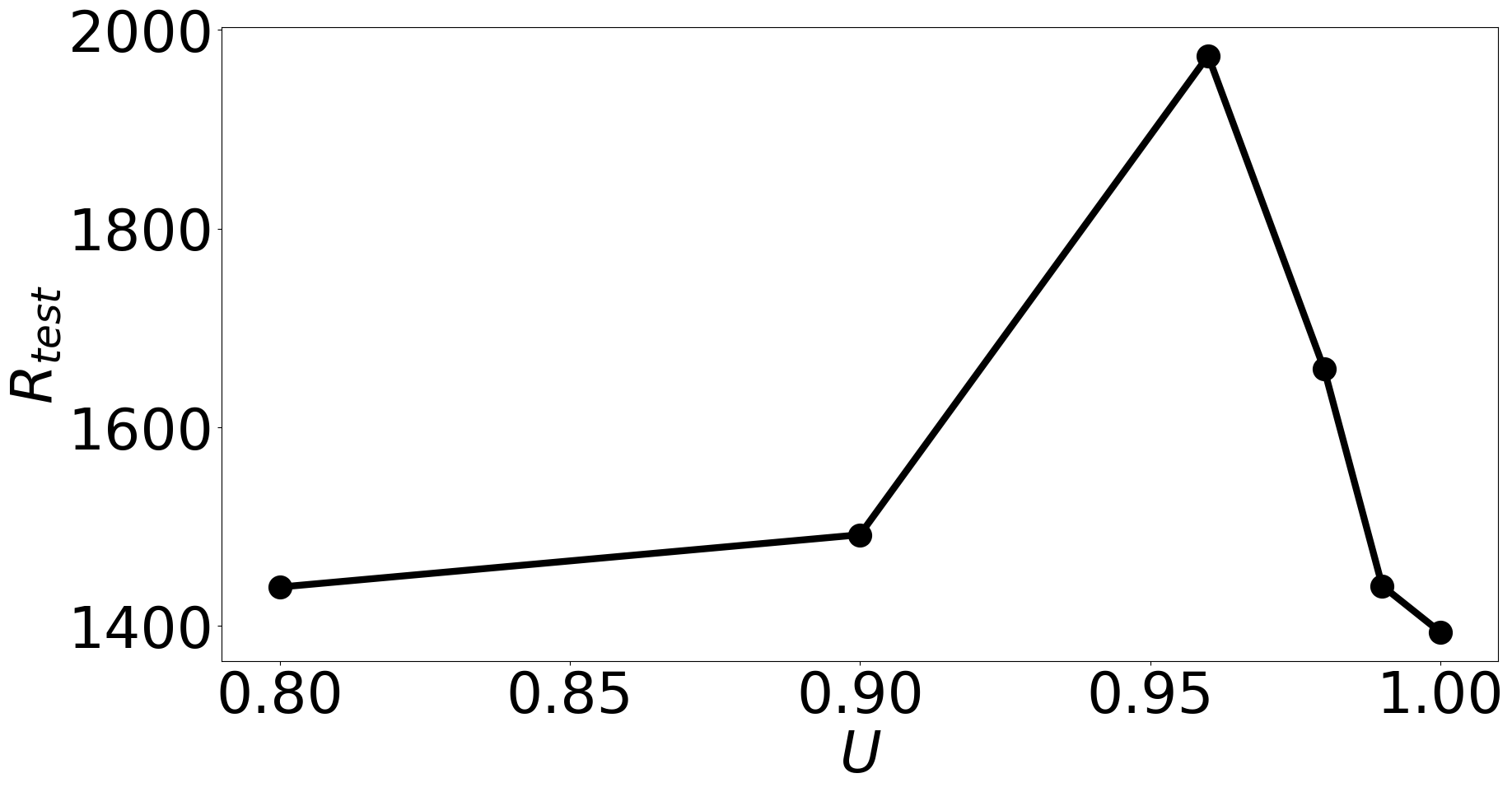}
\caption{Testing rewards with respect to ratio uncertainty levels.} 
\label{fig:test}
\vspace{-3mm}
\end{figure}

Figure~\ref{fig:test} shows the overall performance of the six schemes evaluated by a testing reward. Again, PPO is equivalent to PPO-UE when $U=1$. Thus, we can rank the aforementioned six schemes with respect to different ratio uncertainty levels $U$. The overall performance order is: $\textrm{PPO-UE}_{0.96}\geq\textrm{PPO-UE}_{0.98}\geq\textrm{PPO-UE}_{0.9}\geq\textrm{PPO-UE}_{0.8}\approx\textrm{PPO-UE}_{0.99}\geq\textrm{PPO}$. It is noticeable that PPO performs the worst in the testing phase, which deviates from the performance order in the training phase. This is because PPO is trained with the largest ratio uncertainty level $U$, which destabilizes the training results. Due to the destabilized training, the learned policy is not stable enough in a generalized testing environment. $\textrm{PPO-UE}_{0.98}$ and $\textrm{PPO-UE}_{0.99}$ also have the same issue as their performance rankings drop from the training phase to the testing phase. Overall, the testing reward increases to its optimal at $U=0.96$, and after then drops. This is because of the trade-off between exploration and exploitation. In general, the schemes with low ratio uncertainty levels cannot explore well.  They converge to sub-optimal policies while schemes with high ratio uncertainty levels cannot exploit well. They also perform well with high variance.

\subsection{Posterior Ratio Uncertainty Level}

\begin{figure}[ht]
\centering
\includegraphics[width=0.45\textwidth]{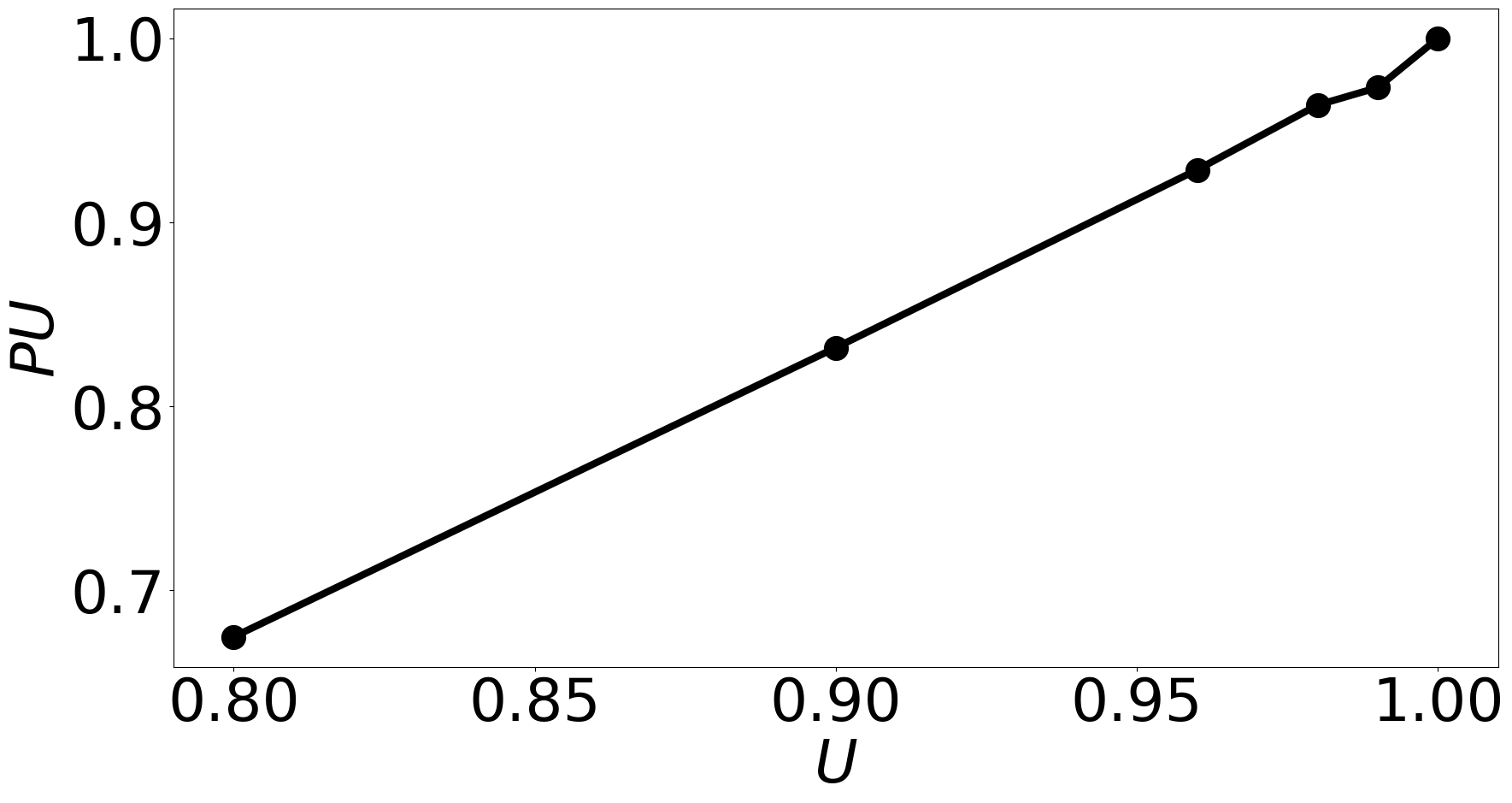}
\caption{Posterior
ratio uncertainty levels with respect to ratio uncertainty levels.} 
\label{fig:pu}
\vspace{-3mm}
\end{figure}

Figure~\ref{fig:pu} shows the posterior ratio uncertainty levels with respect to varying ratio uncertainty levels. We observe that the posterior ratio uncertainty level is positively related to the ratio uncertainty level. Furthermore, they are linearly related. This means we can effectively control the sampling process using predefined ratio uncertainty levels.

\section{Conclusion \& Future Work} \label{sec:conclusion}

In this work, we proposed PPO-UE, a PPO variant with enhanced sampling technique based on a well-defined uncertainty metric, i.e., ratio uncertainty level. The ratio uncertainty level provides a simple but efficient approach to balance exploration and exploitation in the policy training process. By incorporating this technique, PPO-UE can achieve faster convergence and better performance than the PPO baseline with an adaptive uncertainty level.  However, this result is only limited to the given environment considered in this work. As our future work, we will delve into how to generalize our approach to multiple agent cases under more complex environments.

\section{Acknowledgment} \label{sec:acknowledgment}
This work is partly supported by the Army Research Office under Grant Contract Number W91NF-20-2-0140 and NSF under Grant Numbers 2107449, 2107450, and 2107451.
\bibliography{ref}

\end{document}